\documentclass{article}

\usepackage[preprint]{neurips_2025}

\usepackage{subcaption}     
\usepackage[ruled,vlined,linesnumbered]{algorithm2e}
\usepackage[utf8]{inputenc} 
\usepackage[T1]{fontenc}    
\usepackage{hyperref}       
\usepackage{url}            
\usepackage{booktabs}       
\usepackage{amsfonts}       
\usepackage{nicefrac}       
\usepackage{microtype}      
\usepackage{xcolor}         

\usepackage{graphicx}
\usepackage{tabularx}
\usepackage{amsmath}

\usepackage{booktabs}
\usepackage{caption}
\usepackage{graphicx}
\usepackage{multirow}
\usepackage{tabularx}
\usepackage{booktabs} 
\usepackage{array}    
\usepackage{tcolorbox}
\usepackage{tikz}
\usepackage{wrapfig,lipsum}
\usepackage[export]{adjustbox}
\usepackage{wrapfig}

\title{MoE-GPS: \underline{G}uidlines for \underline{P}rediction \underline{S}trategy for Dynamic Expert Duplication in MoE Load Balancing}

\author{%
  Haiyue Ma \\
  Princeton University\\
  \texttt{hm1@princeton.edu} \\
  \And
  Zhixu Du \\
  Duke University \\
  \texttt{zhixu.du@duke.edu} \\
  \AND
  Yiran Chen \\
  Duke University \\
  \texttt{yiran.chen@duke.edu} \\
}

\begin{document}

\maketitle

\begin{abstract}
In multi-GPU Mixture-of-Experts (MoE) network, experts are distributed across different GPUs, which creates load imbalance as each expert processes different number of tokens. Recent works improve MoE inference load balance by dynamically duplicating popular experts to more GPUs to process excessive tokens, which requires predicting the distribution before routing. In this paper, we discuss the tradeoff of prediction strategies, accuracies, overhead, and end-to-end system performance. We propose \textbf{MoE-GPS}, a framework that guides the selection of the optimal predictor design under various system configurations, by quantifying the performance impact to system-level model runtime. Specifically, we advocate for \textbf{Distribution-Only Prediction}, a prediction strategy that only predicts overall token distribution which significantly reduces overhead compared to the traditional Token-to-Expert Prediction. On Mixtral 8×7B MMLU dataset, MoE-GPS suggests Distribution-Only Prediction which improves end-to-end inference performance by more than $23\%$ compared with Token-to-Expert Prediction.  
\end{abstract}

\section{Introduction}

\begin{wrapfigure}{r}{0.5\textwidth}
\includegraphics[width=0.9\linewidth]{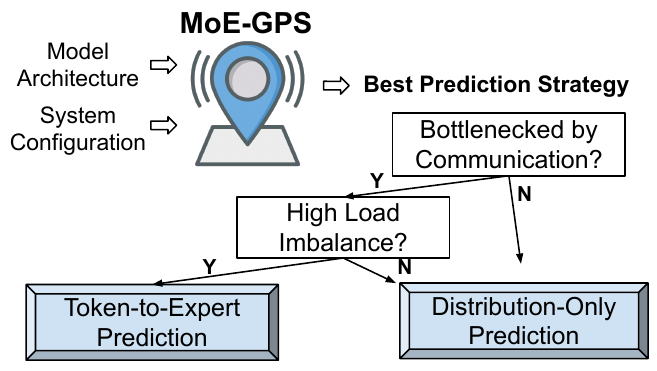} 
\caption{MoE-GPS guidelines for selecting optimal expert prediction strategies that minimizes end-to-end inference latency based on model and hardware characteristics. 
}
\label{fig:figure1}
\vspace{-1em}
\end{wrapfigure}

Mixture-of-Experts (MoE)~\cite{jacobs1991adaptive, jordan1994hierarchical, fedus2022switch, shazeer2017} models reduce the computation of Large Language Models (LLMs) by activating only a subset of experts per token. In large multi-GPU datacenters, Expert Parallelism (EP)~\cite{lepikhin2020gshard} is typically used for Feed Forward Network (FFN) layers, where each GPU only hosts one or few experts. However, due to skewed token-to-expert distribution~\cite{rajbhandari2022deepspeed, cong2024prediction}, this approach often results in load imbalance, underutilized resources, and increased end-to-end latency. The growing number of experts in modern LLM models exaggerates the imbalance~\cite{liu2024deepseek}.

While both training and inference can have imbalance, existing works proactively balance token distribution in training, such as using auxiliary loss~\cite{wang2024auxiliary} which can achieve near-perfect balance. For inference, expert-to-token mappings are fixed and imbalance is unavoidable. A popular approach to mitigate load imbalance is \textit{expert duplication}~\cite{cong2024prediction, wang2023prophet, he2022fastermoe, nie2023flexmoe}, where heavily used experts are replicated across GPUs to distribute tokens more evenly. Since the token distribution changes over time, dynamic duplication is often needed, which requires a predictor for the distribution ahead of the routing stage. Higher prediction accuracy improves load balancing but also incurs greater overhead, creating a complex tradeoff between predictor complexity, accuracy, and system performance. Moreover, the effectiveness of each prediction strategy depends on multiple factors, such as workload characteristics, hardware communication schemes, and the degree of token imbalance.



Despite its importance, there is currently no systematic method to model the runtime implications of MoE load imbalance, and to help choose the best predictor for different workload and system setups. We present \textbf{MoE-GPS}, a framework that simulates end-to-end MoE inference performance with imbalance, and guides the selection of expert prediction strategies that yield the shortest runtime. Built on top of an architectural simulator, LLMCompass~\cite{zhang2024llmcompass}, MoE-GPS models the runtime tradeoffs among prediction strategies, accuracy, and overhead. Given an arbitrary model architecture and hardware setup, MoE-GPS identifies the strategy that delivers the best system performance.

With insights from MoE-GPS, we advocate for \textbf{Distribution-Only Prediction} strategy, which only predicts the coarse-grained token distribution across experts instead of exact token-to-expert mappings (Token-to-Expert Prediction). This lightweight approach is particularly effective when communication is not a bottleneck, because it reduces prediction complexity and still improves compute load balancing. Exact token-level prediction optimizes both computation and communication at the cost of higher overhead, which becomes more favorable when communication cost dominates. In addition, we observe that Distribution-Only Prediction performs better with more balanced workloads. 
These findings are summarized as design guidelines in Figure~\ref{fig:figure1} and further elaborated in Section~\ref{sec:results} and~\ref{sec:discussion_and_limitation}. 

Our contribution can be summarized as follows:
\begin{itemize}
    \item We propose \textbf{MoE-GPS}, a system performance simulation framework that selects the optimal expert prediction strategy to minimize end-to-end inference latency, given a model and hardware setup. 
    \item We identify and validate the effectiveness of \textbf{Distribution-Only Prediction} as a lightweight alternative to Token-to-Expert Prediction, offering better scalability and efficiency under varying system bottlenecks. For example, Distribution-Only Prediction introduces more than $23\%$ system-level performance improvements on Mixtral 8×7B MMLU dataset compared to Token-to-Expert Prediction.
\end{itemize}

Our work provides a practical tool and actionable insights for designing MoE predictors that improve inference speed across various workloads and hardware configurations.
\section{Background}

\begin{figure*}[t]
    \centering
    \includegraphics[width=\textwidth]{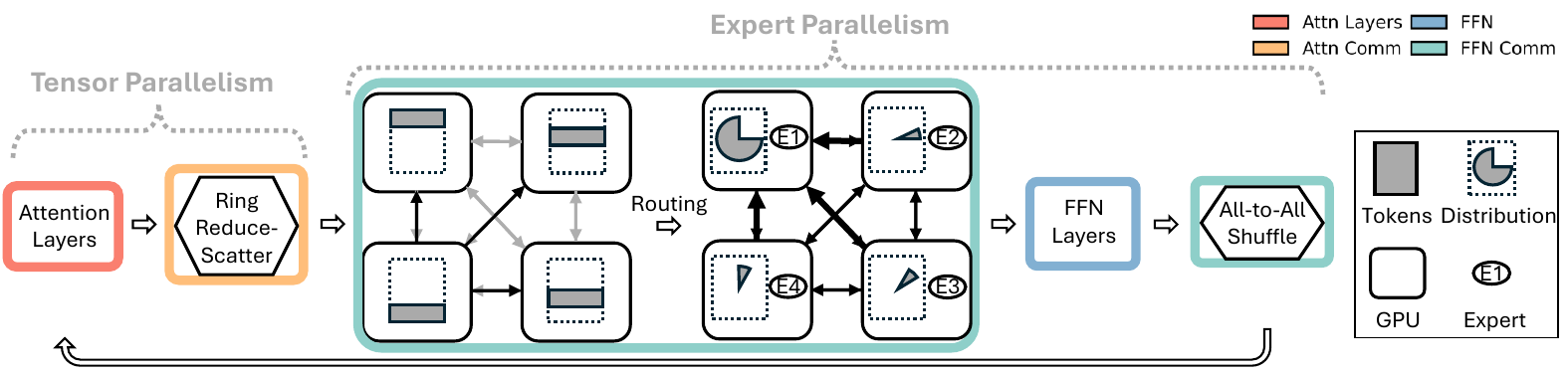}
    \caption{Overview of a typical Mixture-of-Expert inference based on a Transformer network, with four-way Tensor Parallelism and Expert Parallelism. Four Experts (denoted E1-E4) each resides in one GPU. Routing assigns each token to its expert, which creates load imbalance for both compute and communication, if the token-to-expert distribution is skewed.} \label{fig:e2e_illustration}
\end{figure*}

\paragraph{Load Imbalance in Inference Prefill Is Critical.}
Imbalance can occur in both training and inference in a MoE network. Current approaches mitigates training imbalance by proactively routing tokens to less popular experts~\cite{wang2024auxiliary}. In inference, token-to-expert mapping is fixed. Discarding or re-routing excessive tokens on popular experts may improve balance at the cost of model accuracy.

In this paper, we focus on \textbf{inference prefill}, and do not make changes to the original token-to-expert mapping. The decode stage has much less tokens being processed at the same time and is latency-critical, and load balancing is a secondary issue.

Figure~\ref{fig:e2e_illustration} shows the overall flow of a Transformer Block in a Mixture-of-Expert \textbf{inference} workload. For illustration purposes, we assume a four-GPU system, four experts (denoted E1-E4) each residing on one GPU, and each token is routed to its top-1 experts~\cite{fedus2022switch}. The general concepts apply to arbitrary number of devices, experts, and tokens with top-K experts~\cite{jiang2024mixtral, zhu2024llama}. 

\paragraph{Expert Parallelism Leads to Imbalance.}
The parallelism scheme we use is Tensor Parallelism (TP) for the Attention layers and Expert Parallelism (EP) for the FFN layers~\cite{shoeybi2019megatron}. Using TP for the Attention layers excludes the need to duplicate KV cache for adjacent tokens when the sequence is split to different GPUs. We used Ring All Reduce~\cite{ring_all_reduce} to optimize the communication runtime after TP. Using EP for the FFN layers has two benefits over TP: first, it reduces communication latency by the factor of number of devices; second, it avoids splitting the weights in each expert to narrower matrices, which can lower Tensor Core utilization and increase compute time. The introduction of EP assigns experts to certain GPUs and creates load imbalance. Industry practices suggest more advanced hybrid parallelism (TP+EP)~\cite{moe_package} when necessary. For simplicity, this work assumes TP-only Attention and EP-only FFN, and the insights are generalizable.


\paragraph{Quantifying Imbalance.}
We assume that Expert 1 is the most popular expert which takes 75\% of all tokens as shown in Figure~\ref{fig:e2e_illustration}. FFN computation and communication runtime will be dominated by GPU 1 which has Expert 1. To quantify the load imbalance of each GPU, we introduced a parameter called "\textbf{skewness}", the number of tokens in the most popular expert divided by the average number of tokens per expert if the workload is evenly distributed:
\[
\text{skewness} = \frac{\# \text{ of tokens in the most popular expert}}{\# \text{ of average tokens per experts}}
\]

Figure~\ref{fig:e2e_illustration}'s workload skewness is 3 since Expert 1 has 75\% of the tokens but the average token per expert is 25\%. 

\paragraph{Performance Impacts of Load Imbalance.}
Skewness only impacts the runtime of FFN compute and communication since tokens are routed to the experts for FFN. FFN computation typically fully utilizes GPU resources, and the bottleneck FFN runtime is increased by a factor of the skewness. 

Routing incurs inter-device communication. With a fully connected multi-GPU topology, for a perfectly balanced distribution (skewness = 1) each GPU needs to move $(N-1)/N$ of its existing tokens to other GPUs (N = number of GPUs), assuming the token placement is completely random after scatter. This yields $(N-1)/N^2$ tokens to move per GPU. However, for a workload with skewed distribution, the GPU with the most tokens will be receiving more tokens than other GPUs, and thus has longer communication latency. The overall communication time, which is bottlenecked by the GPU with the most popular expert, is scaled by skewness: $(N-1) \cdot \text{skewness}/N^2$. The same amount of communication happens for All-to-All shuffle after the FFN layers.


\paragraph{Current Solutions.}
In a multi-GPU datacenter setting, many existing solutions have proposed expert duplication - mirroring popular experts in other GPUs - to offload excessive token processing. MoE-Prediction~\cite{cong2024prediction} observes highly skewed token-to-expert distribution and proposes predicting such distribution to guide expert placement. Prophet~\cite{wang2023prophet}, FlexMoE~\cite{nie2023flexmoe}, SE-MoE~\cite{shen2022se}, and FasterMoE~\cite{he2022fastermoe, he2021fastmoe} propose specific strategies for dynamic expert duplication. 

Other techniques focus on reducing the distribution imbalance in training by imposing constraints, such as auxiliary load balancing~\cite{wang2024auxiliary} and expert biasing~\cite{liu2024deepseek}. In a different setting with fewer GPUs and the GPU memory capacity is limited for storing experts, token-to-expert mappings are also predicted to facilitate expert offloading to CPU~\cite{du2024sida}. Others proposes different solutions to serve as the token-to-expert mapping, such as expert buffer that stores the hottest experts~\cite{huang2023towards} and expert activation correlation model~\cite{yi2023edgemoe}. These solutions are orthogonal to the discussion of this paper.
\section{Methodology}
This section discusses our approach to modeling and analyzing expert prediction strategies in MoE inference. We describe the integration of expert duplication into the MoE model architecture, and the two different prediction strategies (Distribution-Only and Token-to-Expert). 
We also show normalized system performance obtained from a performance simulator, LLMCompass~\cite{zhang2024llmcompass}, to illustrate high-level trends of the runtime implications of expert duplication. We defer detailed analyses of LLMCompass's simluated performance to Section~\ref{sec:results}.

\begin{figure*}[t]
    \centering
    \includegraphics[width=0.99\textwidth]{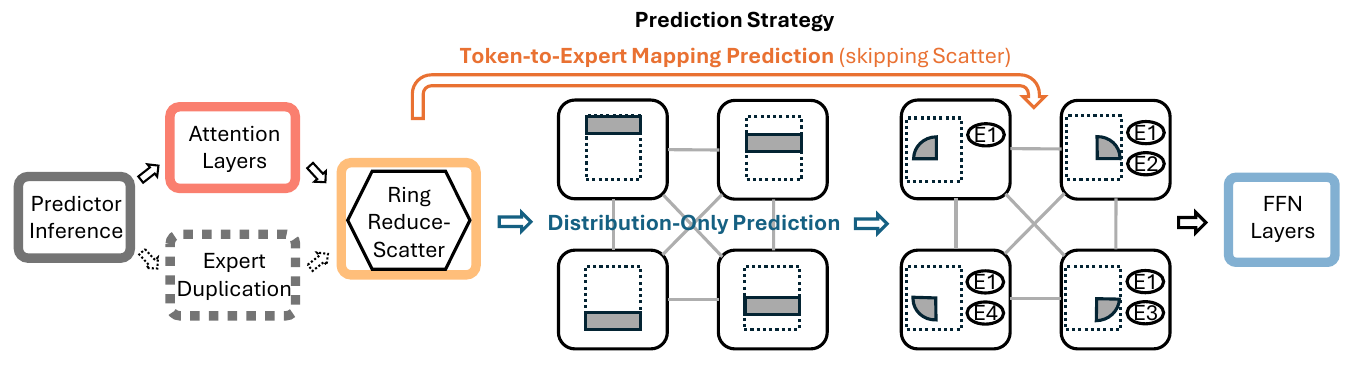}
    \caption{Integration of expert duplication and different prediction strategies into the MoE model architecture for a single layer. Token-to-Expert Prediction eliminates communication by directly routing tokens to their predicted GPUs, while Distribution-Only Prediction assumes random token distribution during scattering.}   \label{fig:prediction_illustration}
\end{figure*}

\subsection{Approach to Expert Duplication}
We integrate dynamic expert duplication into the MoE model by inserting a pre-trained predictor before Attention in each layer. For a chosen frequency, the predictor does inference with current batched inputs to predict distribution and guide the expert placement. Since dynamic duplication incurs extra inter-GPU communication for moving experts, existing works propose different intervals for prediction and moving, from every single batch~\cite{he2021fastmoe, wang2023prophet} to every 10 minutes~\cite{liu2024deepseek}, to balance overhead and effectiveness.

In this study, we assume single-batch prediction and placement frequency. We show calculation in Section~\ref{sec:discussion_and_limitation} that the placement latency can be hidden within the Attention layers which is compute-bound with moderate batch sizes. Our simulator can be configured to model different frequencies of prediction and placement, by averaging out the overhead to multiple batches.

Figure~\ref{fig:prediction_illustration} illustrates how expert duplication helps balance token workloads. 
In our example, Expert 1 has the most tokens, so it is replicated across multiple GPUs to evenly distribute the load. In general, given arbitrary token-to-expert mappings, experts can be duplicated to achieve per-GPU balance by iteratively shifting experts from overloaded to underloaded devices: keep duplicating the experts on GPUs with $>1/N$ tokens to GPUs with $<1/N$ tokens until all GPUs process the same amount of tokens. A detailed algorithm can be found in Algorithm~\ref{algo:expert_duplication}.

\begin{algorithm}[t]
\caption{Expert Duplication in MoE Load Balancing}
\label{algo:expert_duplication}
\KwIn{%
Token–expert map $f:\{1,\dots,T\}\!\to\!\{1,\dots,E\}$; Per-GPU memory capacities $M\in\mathbb{N}^G$;\\
Initial placement $\mathcal{P}\subseteq\{1,\dots,E\}\!\times\!\{1,\dots,G\}$; 
Maximum copies per expert $C_{\max}$.}
\KwOut{%
Balanced placement $\mathcal{P}$ and dispatch $d:\{1,\dots,T\}\!\to\!\{1,\dots,G\}$.}
\BlankLine
Assign each token $t$ to any GPU $g$ that hosts its expert: $d(t)\leftarrow\min\{g\mid(f(t),g)\in\mathcal{P}\}$\;
$L_g\leftarrow|\{t\mid d(t)=g\}|$ for all $g\in\{1,\dots,G\}$;

\While{$\max_g L_g-\min_g L_g>1$}{
    $g_h\leftarrow\arg\max_g L_g$;\quad
    $g_c\leftarrow\arg\min_g L_g$\;
    $\Delta\leftarrow\left\lceil\frac{L_{g_h}-L_{g_c}}{2}\right\rceil$\;
    $e^\star\leftarrow\arg\max_{e\in\mathcal{E}(g_h)}|\{t\mid d(t)=g_h\land f(t)=e\}|$\;
    \If{$(e^\star,g_c)\notin\mathcal{P}$ \textbf{and} copies$(e^\star)<C_{\max}$ \textbf{and} params$(e^\star)\le M_{g_c}$}{
        copy weights of $e^\star$ on $g_c$;\quad
        $\mathcal{P}\leftarrow\mathcal{P}\cup\{(e^\star,g_c)\}$\;
    }
    Reassign the first $\Delta$ tokens $\{t\mid d(t)=g_h\land f(t)=e^\star\}$ to $g_c$ by setting $d(t)\leftarrow g_c$\;
    Update $L_{g_h}$ and $L_{g_c}$\;
}
\Return{$\mathcal{P},d$}\;
\end{algorithm}

\subsection{Prediction Strategies and Tradeoffs}
\label{sec:pred_strategies}
We explore two prediction strategies with distinct tradeoffs: \textit{Distribution-Only Prediction}, which estimates static, aggregate expert usage; and \textit{Token-to-Expert Prediction}, which targets exact token-level routing. The former has lower overhead and complexity and targets at compute imbalance, while the latter can reduce both compute and communication costs at the cost of high overhead.

\subsubsection{Distribution-Only Prediction}

Distribution-only prediction estimates the proportion of tokens routed to each expert (e.g., Expert 1 gets 75\% of tokens) without specifying which tokens. This enables balanced compute across GPUs but does not reduce communication costs, as tokens are still randomly scattered post-ring all-reduce.

We model the expert activation distribution in each layer of the MoE model using a multinomial distribution and estimate its parameters via Maximum Likelihood Estimation (MLE). The multinomial distribution is commonly used to model counts of outcomes across discrete classes, in our case the selection of experts within a layer. MLE selects the distribution that the observed data has the maximum likelihood to fit into.


\begin{wraptable}{r}{0.5\textwidth}
  \centering
  \vspace{-10pt}
  \caption{Impact of skewness on expert distribution estimation and system performance. Higher skewness leads to higher error rate, indicating reduced estimation accuracy and degraded performance.}
  \label{tab:dist_only_skew_vs_perf}
  \begin{tabular}{lcc}
    \toprule
    \textbf{Dataset} & \textbf{Skewness} & \textbf{Error rate ($\%$)} \\
    \midrule
    MMLU           & 1.39     & 1.80 \\
    Alpaca Eval    & 1.40     & 0.98\\
    SST2           & 1.99     & 16.00 \\
    \bottomrule
  \end{tabular}
  \vspace{-10pt}
\end{wraptable}

Formally, let $p_i^l$ denote the probability of selecting expert $i$ in layer $l$. Assuming that each token's expert selection is an independent and identically distributed (i.i.d.) sample from this multinomial distribution, the MLE of $p_i^l$ is given by:

\begin{equation}
    \hat{p}_i^l = \frac{n_i^l}{N},
\end{equation}

where $N$ is the total number of tokens observed at layer $l$, and $n_i^l$ is the number of tokens that activated expert $i$. When the training data come as batches, the estimation becomes a moving average. We note that expert selection is primarily governed by local token-level features and routing mechanisms. Hence, modeling activations as independent draws to mimick per-batch distribution provides a reasonable approximation while significantly simplifying analysis. We provide detailed derivation in Appendix~\ref{sec:mle}.

We experiment the Distribution-Only Prediction on three datasets: MMLU~\cite{hendrycks2020measuring}, Alpaca Eval~\cite{dubois2024length}, and SST2~\cite{socher-etal-2013-recursive}, on the Mixtral 8×7B~\cite{jiang2024mixtral} MoE model. For each batch, we set sequence length to 512 and measure average skewness across batches: 1.388, 1.402, and 1.990 respectively. We measure the averaged \textbf{error rate} over layers between our estimation of the probability on the trainset and empirical probability on the testset. We define error rate as $$\frac{\vert\hat{p} - p\vert}{1/\text{\# of experts}}.$$ Higher error rate indicates less accurate estimation. For datasets that do not have a dedicated test split, we use train test split to randomly partition the trainset with $80\%$ training samples and $20\%$ test samples. 

We evaluate the normalized end-to-end system performance across all datasets given the model size (Mixtral 8×7B), network parameters (batch size = 1, sequence length = 512), and hardware configurations (four A100 GPUs, fully connected with NVLINK). Performance is simulated with an augmented version of LLMCompass~\cite{zhang2024llmcompass}. The error rate is used to scale the runtime of the GPU that processes the most tokens, which will be detailed in Section~\ref{sec:err_rate}.


Table~\ref{tab:dist_only_skew_vs_perf} shows that lower error rate in expert distribution (more accurate estimation) leads to improved system performance. Notably, higher skewness results in larger error rates and degraded normalized performance. This is primarily because greater skewness leads to underutilized experts that receive fewer tokens. The reduced token count increases estimation error for these experts, which disproportionately contributes to the overall error rate.

\subsubsection{Token-to-Expert Prediction}
Token-to-Expert Prediction targets exact routing of each token to its expert's GPU. With predicted mapping, we can send each token to the GPU that has its experts directly. This strategy skips the scatter phase in communication after the ring all-reduce, and saves both FFN compute and communication runtime, as shown in Figure~\ref{fig:prediction_illustration}. 

We formulate the expert selection task as a classification problem, where the goal is to predict the activated expert for each token in the batch that will be processed by the MoE model. We explore three types of models: a simple probability-based model, a conditional probability model, and neural network-based predictors.
\begin{itemize}
    \item \textit{Probability Model.} We always assign the expert with the highest global frequency observed in the training data, treating all tokens identically regardless of their identity or position.
    \item \textit{Conditional Probability Model.} This model improves by conditioning on either the token index or its position index. For each token (or position), we select the expert that appears the most frequently in the training data for that specific token index or position index.
    \item \textit{Neural Networks.} We train neural models using pairs of token embeddings and their corresponding expert activations. The models are optimized using cross-entropy loss and the Adam optimizer~\cite{kingma2014adam} and trained to converge. All samples are padded to sequence length 512. We experiment with both simple Feed-Forward Networks (FFNs) and Long Short-Term Memory networks (LSTMs)~\cite{hochreiter1997long}. The detailed architectures of these networks are provided in Appendix~\ref{sec:arch}.
\end{itemize}

\begin{figure}[t]
    \centering

    \begin{subfigure}[b]{0.5\textwidth}
        \includegraphics[width=\linewidth]{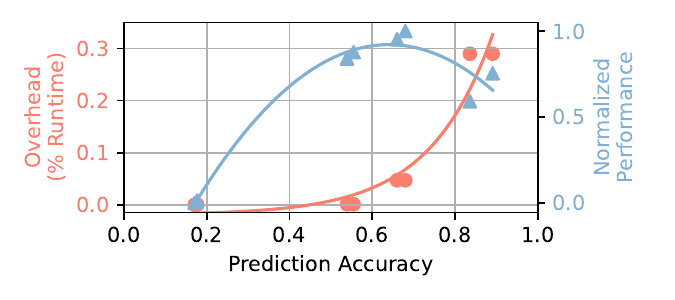}
        \caption{Mixtral 8×7B (skew = 1.4).}
        \label{fig:precision_vs_overhead_curve_cc_Mixtral_skew1.4}
    \end{subfigure}%
    \hfill
    \begin{subfigure}[b]{0.5\textwidth}
        \includegraphics[width=\linewidth]{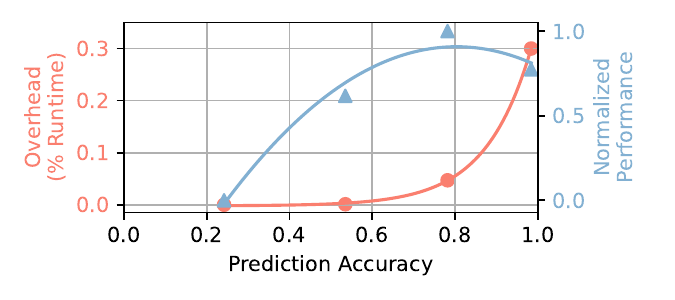}
        \caption{Mixtral 8×7B (skew = 2.0).}
        \label{fig:precision_vs_overhead_curve_sst_Mixtral_skew2.0}
    \end{subfigure}

    \caption{Trade-off between prediction accuracy and end-to-end system performance for Token-to-Expert Prediction.
Increasing prediction accuracy improves expert placement but also raises runtime overhead. Although higher accuracy results in better initial expert placement, excessive overhead can reduce overall gains. In higher-skewness settings (right), accurate predictions are easier to achieve, shifting the optimal performance point toward higher accuracy.}
    \label{fig:precision_vs_overhead_curve_mixtral}
\end{figure}

We investigate Token-to-Expert Prediction on the same three datasets, i.e., MMLU, Alpaca Eval, and SST2. The results are shown in Figure~\ref{fig:precision_vs_overhead_curve_cc_Mixtral_skew1.4} with the MMLU and Alpaca Eval data set (shown together because they have similar skewness) and Figure~\ref{fig:precision_vs_overhead_curve_sst_Mixtral_skew2.0} with the SST2 dataset, for Mixtral 8X7B. The $x$-axis refers to the prediction accuracy for the predictor, each point representing a different predictor being used. The overhead $y$-axis shows the percentage of the prediction overhead of the total runtime for the MoE model, measured on A100 GPUs. The normalized performance $y$-axis is the simulated end-to-end system performance by LLMCompass, using four A100 GPUs fully connected with NVLink, and batch size = 1 and sequence length = 512. The runtime includes the overhead of each prediction strategy. We use exponential functions to fit the accuracy to overhead curves and polynomial functions to fit the accuracy to performance curves.

Higher prediction accuracy can result in better initial expert placement leading to better system performance. However, higher accuracy also incurs larger overhead, which degrades the end-to-end performance. The figure demonstrates a trade-off that it is not always the case that predictors with higher accuracy are better. 
Figure~\ref{fig:precision_vs_overhead_curve_cc_Mixtral_skew1.4} and Figure~\ref{fig:precision_vs_overhead_curve_sst_Mixtral_skew2.0} also demonstrates results of different skewness. Note that, for scenarios with higher skewness, it costs less for the predictor to acquire higher accuracy. Consequently, the sweet point of the system performance moves towards higher accuracy end.

\subsection{Modeling the Effect of Prediction Error}
\label{sec:err_rate}
\begin{wrapfigure}{r}{0.45\textwidth}
\centering
\includegraphics[width=0.7\linewidth]{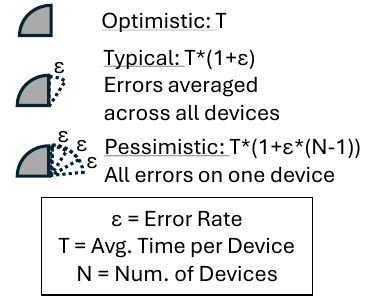} 
\caption{Modeling the impact of prediction errors on end-to-end system runtime.
Three scenarios for the same prediction error rate $\epsilon$: (1) Optimistic—errors do not affect load balancing; (2) Typical—errors are evenly distributed across devices, leading to moderate slowdown; (3) Pessimistic—errors concentrate on one device, causing worst-case load imbalance.}
\vspace{-2em}
\label{fig:prediction_accuracy}
\end{wrapfigure}

We model the system-level effects of \textbf{imperfect prediction} by analyzing how different error distributions affect FFN computation load balance and communication cost. 

First, we model FFN compute's runtime with prediction error rate $\epsilon$, which is $1-\text{accuracy}$ for Token-to-Expert Prediction, and averaged $L_1\ \text{distance}$ over number of experts for Distribution-Only Prediction.

Figure~\ref{fig:prediction_accuracy} shows three potential outcomes for the same prediction accuracy for FFN compute's runtime. Assume that $\epsilon=0.1$:
\begin{itemize}
    \item \textit{Optimistic}: Errors still result in perfect load balancing; e.g. predicting 85\% of tokens instead of 75\% for Expert 1 in Figure~\ref{fig:e2e_illustration}.
    \item \textit{Typical}: Errors are uniformly distributed across GPUs, leading to moderate imbalance. The most loaded GPU processes up to $(1 + \epsilon) \cdot \text{avg\_tokens}$. This is the default model used in our runtime simulations.
    \item \textit{Pessimistic}: All errors occur on a single GPU, leading to worst-case imbalance where the bottleneck GPU handles up to $N \cdot (1 + \epsilon) \cdot \text{avg\_tokens}$. While unlikely to happen, this scenario represents an upper bound on performance degradation.
\end{itemize}


We apply the same typical-case assumption when modeling communication overhead under Expert Parallelism (EP) for Token-to-Expert Prediction. Unlike compute, however, communication costs always increase with prediction errors, as misrouted tokens inevitably trigger additional inter-GPU data transfers. Optimistic cases do not exist in this context.

\subsection{Network Performance Simulation}
We conduct performance evaluations using an extended version of LLMCompass~\cite{zhang2024llmcompass}, a block-level simulator for large language model inference, validated with silicon measurements. This simulation-based approach allows us to easily explore design tradeoffs across hardware and software configurations without requiring access to real clusters, which are often costly and impractical for exhaustive testing.

Our simulations focus on a single Transformer layer, modeling all relevant operations including GEMM, communication, and element-wise computations. Since LLMCompass does not yet support FlashAttention~\cite{dao2022flashattention}, our Attention layer runtimes are conservatively overestimated.

We augmented LLMCompass with:
\begin{itemize}
\item \textit{MoE and Expert Parallelism (EP)}: We introduced custom modules to model EP-specific communication and FFN workloads.
\item \textit{Mixtral Support}: We added support for Mixtral-style model architectures, integrating existing Grouped Query Attention (GQA)~\cite{ainslie2023gqa} and SwiGLU~\cite{shazeer2020glu} activation implementations. We also implemented Sliding Window attention~\cite{beltagy2020longformer}.
\item \textit{Prediction Strategy Modeling}: We added support for evaluating both Distribution-Only and Token-to-Expert prediction strategies with tunable accuracy and overhead.
\end{itemize}
\section{Results}
\label{sec:results}
In this section, we discuss how to choose the best predictor for overall system performance under different system configurations, workload sizes and token distribution skewnesses. Our goal is to identify when to use \textbf{Distribution-Only Prediction}, which reduces FFN computation load imbalance without communication savings, and when to use \textbf{Token-to-Expert Prediction}, which additionally reduces communication latency at the cost of higher predictor complexity. Additionally, for Token-to-Expert Prediction, we want to find the optimal prediction accuracy to minimize overall runtime.

\begin{figure}[t]
    \centering

    \begin{subfigure}[b]{0.33\textwidth}
        \centering
        \includegraphics[width=\linewidth]{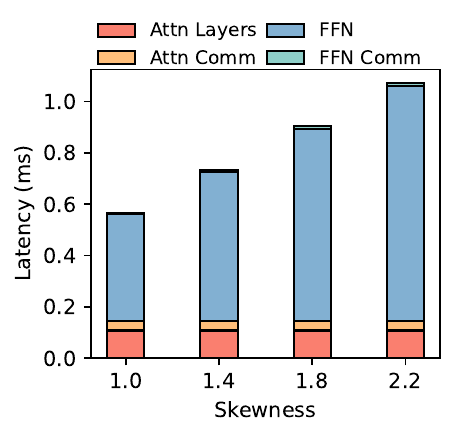}
        \caption{Baseline latency with no prediction (interconnect = NVLink).}
        \label{fig:prefill_skew_mixtral_bs8_sl8192_gpu4_nvlink}
    \end{subfigure}%
    \hfill
    \begin{subfigure}[b]{0.63\textwidth}
        \centering
        \includegraphics[width=\linewidth]{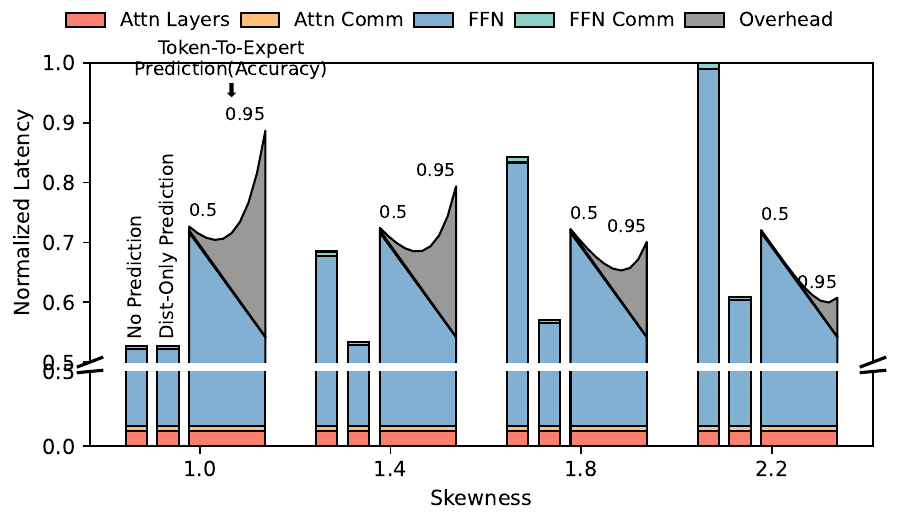}
        \caption{Latency of different prediction strategies and accuracies (interconnect = NVLink).}
        \label{fig:prefill_coarse_prediction_mixtral_bs8_sl8192_gpu4}
    \end{subfigure}

    \vspace{0.7em}  

    \begin{subfigure}[b]{0.33\textwidth}
        \centering
        \includegraphics[width=\linewidth]{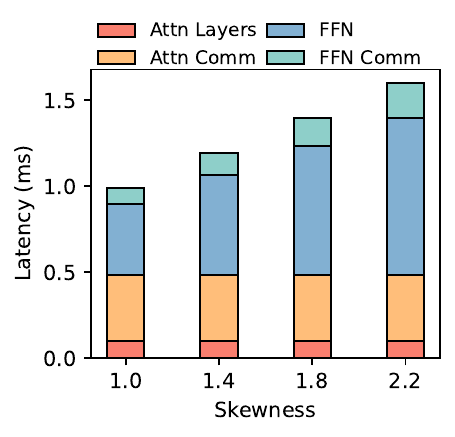}
        \caption{Baseline latency with no prediction (interconnect = PCIe).}
        \label{fig:prefill_skew_mixtral_bs8_sl8192_gpu4_PCIe}
    \end{subfigure}%
    \hfill
    \begin{subfigure}[b]{0.63\textwidth}
        \centering
        \includegraphics[width=\linewidth]{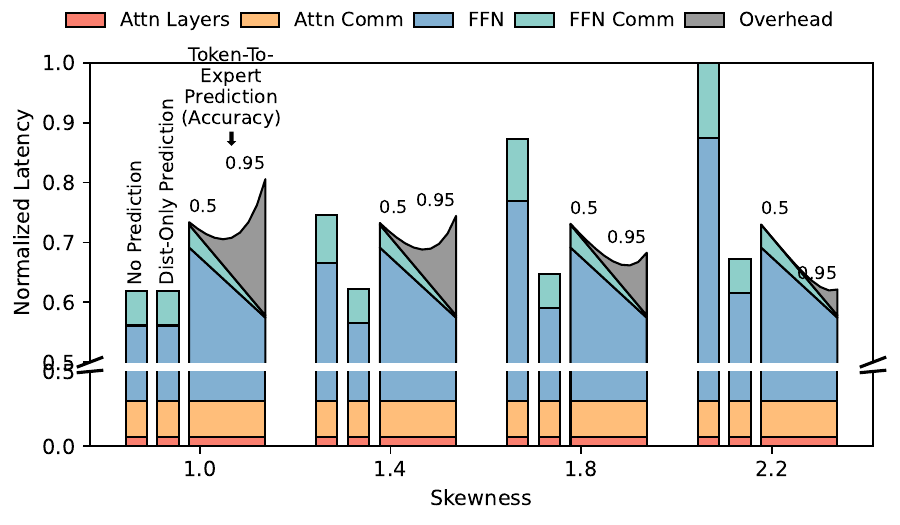}
        \caption{Latency of different prediction strategies and accuracies (interconnect = PCIe).}
        \label{fig:prefill_precise_prediction_mixtral_bs8_sl8192_gpu4}
    \end{subfigure}
    \caption{Simulated prefill latency for a single layer of Mixtral 8×7B under different prediction strategies and interconnect types.
Latency is broken down by component (attention, FFN, communication, overhead) and evaluated across skewness levels on 4 A100 GPUs using NVLink (top) and PCIe (bottom). (a, c) show baseline latencies without prediction; (b, d) show improvements from prediction strategies at varying accuracies. Distribution-Only Prediction reduces FFN compute without overhead, while Token-to-Expert Prediction introduces overhead that trades off with improved load balancing. 
For each skewness, the best preditor has the minimum total latency across strategies and accuracies.
}
    \label{fig:prefill_prediction_mixtral_full_comparison}
\vspace{-1.5em}
\end{figure}

Figure~\ref{fig:prefill_prediction_mixtral_full_comparison} shows the time-to-first-token latency for a single layer of Mixtral 8×7B prefill workload, simulated by LLMCompass. To match the experimental setup described in Section~\ref{sec:pred_strategies}, we set the batch size to 1 and the sequence length to 512. The simulation includes all architectural components of Mixtral, such as a 4K sliding window, Grouped Query Attention, and SwiGLU activation. For system configurations, we modeled four A100s, either connected via NVLink 3.0~\cite{nvlink} (high-end interconnect with 2 TB/s bandwidth) or PCIe 4.0~\cite{pcisig2017pcie4} (low-end interconnect with 32GB/s bandwidth). 

Figures~\ref{fig:prefill_skew_mixtral_bs8_sl8192_gpu4_nvlink} and~\ref{fig:prefill_skew_mixtral_bs8_sl8192_gpu4_PCIe} show baseline latencies without any prediction strategy with NVLink and PCIe interconnect respectively. While FFN and Attention latencies remain constant across configurations, PCIe's limited bandwidth makes communication latency a dominant bottleneck.


Figure~\ref{fig:prefill_coarse_prediction_mixtral_bs8_sl8192_gpu4} and~\ref{fig:prefill_precise_prediction_mixtral_bs8_sl8192_gpu4} show the impact of applying prediction strategies under varying skewness and accuracy. For Token-to-Expert Prediction, we use the fitted curve in Figure~\ref{fig:precision_vs_overhead_curve_mixtral} to model the overhead for each prediction accuracy. For skewness numbers without a matching dataset, we interpolated the overhead from the skewnesses that we measured in Section~\ref{sec:pred_strategies}. 

Within each skewness group, the leftmost bar shows latency without prediction, same as those in Figure~\ref{fig:prefill_skew_mixtral_bs8_sl8192_gpu4_nvlink} and~\ref{fig:prefill_skew_mixtral_bs8_sl8192_gpu4_PCIe}. The second bar shows latency under \textit{Distribution-Only Prediction}. This strategy reduces FFN compute time, though communication time remains unchanged. Since distribution is estimated offline, there is no prediction overhead. As skewness increases, distribution estimation becomes less accurate, slightly diminishing its benefits.

The remaining stacked bars (shown as curves) show latencies for \textit{Token-to-Expert Prediction} across multiple accuracy levels. Each curve exhibits a U-shape: higher accuracy improves load balancing (reducing FFN and communication time), but also increases overhead. The optimal configuration is the one with the lowest total latency, typically at an intermediate accuracy level.

We see that Distribution-Only Prediction performs better in most cases. For example, with skewness = 1.4 (a typical case, close to MMLU dataset), Distribution-Only Prediction achieves 23\% speedup compared to the best configuration of Token-to-Expert Prediction (the bottom of the U-shape). 

Compared to Distribution-Only, Token-to-Expert Prediction incurs significantly higher overhead, especially at low skewness where it requires more complex models. However, it becomes beneficial when 1) skewness is high: accurate predictions are easier to make, reducing overhead, and 2) communication cost is high (e.g., PCIe): saved communication time outweighs added prediction cost.

Across all configurations, FFN latency reductions from prediction are largely skewness-independent. This is because mispredictions are measured relative to a perfectly balanced reference scenario. Only overhead varies with skew as higher skewness makes prediction easier. 

\begin{wrapfigure}{r}{0.5\textwidth}
\includegraphics[width=0.95\linewidth]{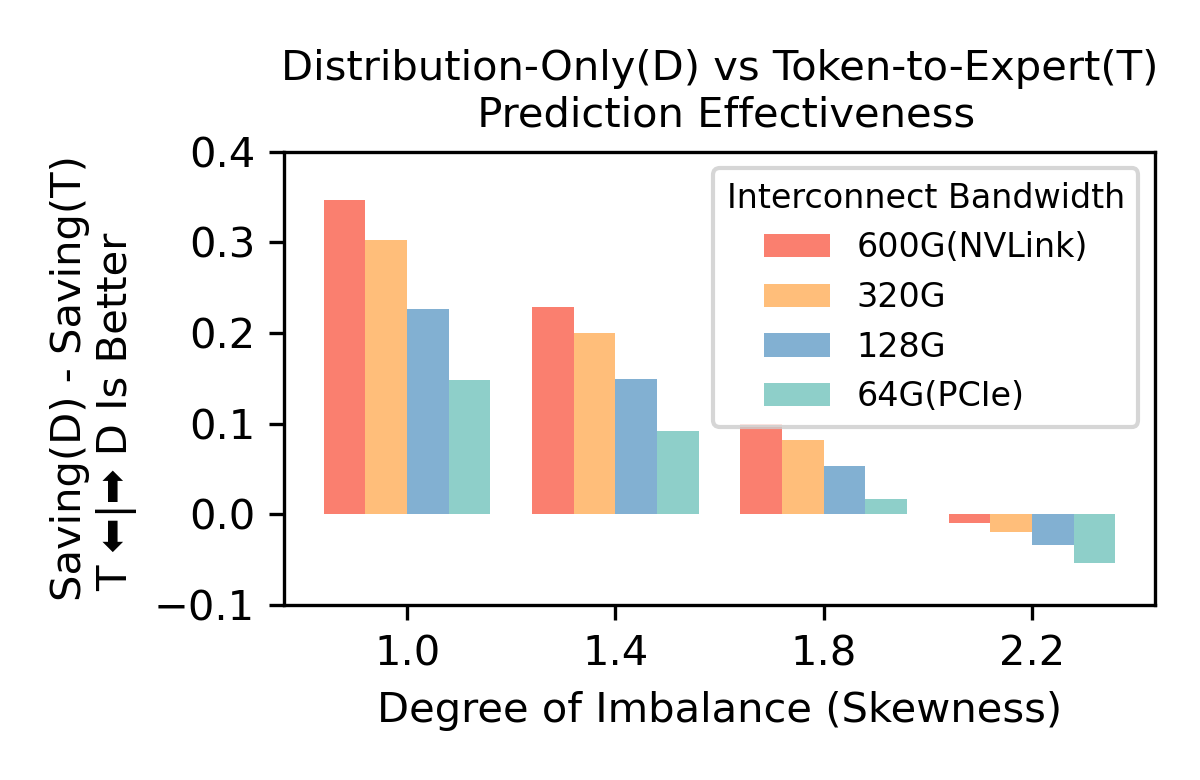} 
\caption{Simulated effectiveness of two prediction strategies' best savings for Mixtral 8X7B on 4 fully-connected A100 with different system interconnect settings.}
\label{fig:skew_bar_plot}
\vspace{-1em}
\end{wrapfigure}

To clearly draw our insights on which prediction strategy is better under different scenarios, we plotted the runtime savings on baseline (no prediction) of the best configuration of Token-to-Expert Prediction (the point of lowest latency for each skewness), and compared that with the savings of Distribution-Only Prediction. Figure~\ref{fig:skew_bar_plot} shows the difference in savings for the two strategies, calculated by $Distribution\text{-}Only\ Prediction\ saving-Token\text{-}to\text{-}Expert\ Prediction\ saving$. Every bar above zero indicates Distribution-Only Prediction outperforms the best configuration of Token-to-Expert Prediction, and every bar below zero indicates otherwise. The 600GB/s and 64GB/s interconnect bandwidth correspond to settings in Figure~\ref{fig:prefill_coarse_prediction_mixtral_bs8_sl8192_gpu4} and ~\ref{fig:prefill_precise_prediction_mixtral_bs8_sl8192_gpu4} respectively. The other two are arbitrary interconnect settings to model cases where a mixture of both interconnects are used (such as across NVLink domains). 

\textbf{Key takeaways:}
\begin{itemize}
\item \textit{Distribution-Only Prediction} is preferred when skewness is low or communication is not a bottleneck. Its low complexity and zero overhead make it a strong strategy.
\item \textit{Token-to-Expert Prediction} becomes more effective as skewness increases, since predictors achieve high accuracy at lower cost. It is particularly beneficial under low-bandwidth interconnects like PCIe, where communication savings dominate.
\end{itemize}

\section{Discussion and Limitation}
\label{sec:discussion_and_limitation}

\paragraph{Generality across model architectures.}
While our results focus on Mixtral 8×7B, the core insights generalize across other MoE-based architectures such as Switch Transformer~\cite{fedus2022switch}, LLaMA variants~\cite{zhu2024llama, touvron2023llama}, and DeepSeek~\cite{liu2024deepseek}. Although these models differ in their expert construction and token routing decisions, we observe consistent system-level behaviors for both \textit{Distribution-Only} and \textit{Token-to-Expert} predictions. Specifically, for token-to-expert mapping, more complex predictors lead to higher accuracy but also higher overhead, which brings negative impact to FFN layers' performance. Supporting experiments for LLaMA and Switch Transformer are provided in Appendix~\ref{sec:llama_and_switch_transformer}.

At the system level, Attention and FFN dominate the cost of MoE inference across models. LLaMA shares Mixtral’s SwiGLU-based FFN but lacks sliding-window Attention; Switch Transformer uses ReLU in FFN and does not use Grouped Query Attention (GQA)\cite{ainslie2023gqa}; DeepSeek introduces Multi-head Latent Attention (MLA)\cite{meng2025transmla} to further optimize Attention. Despite these differences, FFN layers remain a substantial portion of runtime, and expert duplication yields consistent benefits across all. Additionally, scaling model size (e.g., Mixtral 8×7B vs. 8×22B) changes absolute latency but not the qualitative trends or effectiveness of each prediction strategy.

\paragraph{Generality across hardware systems.}
In our study, we assumed fully connected GPUs with the same interconnect bandwidth between each pair. For a larger-scale system with more GPU nodes, different topologies including Mesh, Torus, and Tree topologies can be used. These topology choices will impact specific runtime but are orthogonal to our core insights, and can be modeled by changing the topology implementation.
Further, while we assume full TP for Attention layers and full EP for FFN layers in this study, hybrid parallelism (such as using TP+EP for FFN) have been proven to be useful in certain settings. Support can easily be added by incorporating current frameworks for hybrid parallelism~\cite{qin2025chimera}.

\paragraph{Long sequence lengths.}
The experiments shown in the paper use a sequence length of 512. Longer sequences introduce new tradeoffs, particularly for Token-to-Expert Prediction. For FFN-based predictors, although computation remains parallelizable and overhead manageable, we observe a lower bound on achievable accuracy as sequence length increases. LSTM-based predictors are theoretically sequence-length agnostic in accuracy but suffer from poor parallelism and can hardly scale across different devices. Therefore, under long-sequence workloads, \textit{Distribution-Only Prediction} may become more favorable due to its scalability and low complexity.  

\paragraph{Kernel underutilization at small scale.}
To enable fast evaluation, we use small batch sizes and short sequences, which can expose low-level inefficiencies in kernel execution, such as insufficient overlap between prologue/epilogue and the main MMA loop, which leads to underutilization of Tensor Core FLOPs. Our simulator is throughput-oriented and assumes at least one hardware unit (in this case, compute) is saturated. To validate our findings, we compared simulated results against actual GPU measurements for both model and predictor inference. Although the absolute runtimes differ, the relative overhead between prediction and inference is consistent. Thus, we report and analyze prediction overhead as a ratio to the simulated inference runtime for an accurate estimation.

\paragraph{Expert duplication's communication overhead.}
We also quantify the communication overhead of expert duplication to show that we can hide its latency with the Attention layers. For Mixtral 8×7B, each FP16 expert contains approximately $4{,}096 \times 14{,}336 \times 2 \times 2$ bytes of weight data. Assuming one expert is sent and received per GPU per layer, the transfer takes $\sim$0.1 ms over NVLink 3.0 (2 TB/s bandwidth), smaller than the Attention layer runtime for batch size 1 and sequence length 512. In practice, expert weights are static across inputs, and this duplication can be hidden with Attention computation. Even with PCIe 4.0 (32 GB/s), the duplication latency can be hidden with modest increases in batch size or sequence length (e.g., batch size 16, sequence length 2K), which are still on the low-end of practical inference workload sizes. In more advanced systems, expert prefetching or pipelined movement between layers can further reduce this cost.

\section{Conclusion}
In this work, we presented MoE-GPS, a framework for selecting expert prediction strategies that minimize end-to-end inference latency in Mixture-of-Experts models. By modeling the trade-offs between prediction accuracy, overhead, and system runtime across different hardware and workload configurations, MoE-GPS provides actionable guidance for system designers. We also demonstrated the effectiveness of Distribution-Only Prediction as a lightweight alternative to token-level prediction, particularly in scenarios where load imbalance is low and communication is not a bottleneck.
\newpage

\bibliographystyle{plain}
\bibliography{references}

\newpage
\appendix

\section{MLE for Multinomial Distribution}
\label{sec:mle}
We seek to model the expert activation pattern in a Mixture-of-Experts (MoE) model using a probabilistic approach. Specifically, we are interested in estimating the distribution over expert selections for each MoE layer, based on observed activation frequencies in the training data. To this end, we adopt a multinomial modeling framework and employ Maximum Likelihood Estimation (MLE) to infer the activation probabilities.

\paragraph{Assumptions.}
We assume that each token independently selects an expert from a fixed pool of $K$ experts in a given layer. Let $E = \{e_1, \ldots, e_K\}$ denote the set of experts in a particular MoE layer. For each token routed through the layer, the expert selection is modeled as an i.i.d. draw from a multinomial distribution with parameters $\mathbf{p} = (p_1, \ldots, p_K)$, where $p_i$ is the probability that expert $e_i$ is selected. Naturally, $\sum_{i=1}^K p_i = 1$ and $p_i \geq 0$ for all $i$.

\paragraph{Maximum Likelihood Estimation.}
Given $N$ tokens routed through the layer, let $n_i$ denote the number of tokens that selected expert $e_i$, so that $\sum_{i=1}^K n_i = N$. The likelihood of the observed expert assignments under the multinomial distribution is:

\begin{equation}
\mathcal{L}(\mathbf{p}) = \Pr(n_1, \ldots, n_K \mid \mathbf{p}) = \frac{N!}{n_1! \cdots n_K!} \prod_{i=1}^K p_i^{n_i}.
\end{equation}

To estimate $\mathbf{p}$ via MLE, we maximize the log-likelihood:

\begin{equation}
\log \mathcal{L}(\mathbf{p}) = \log \left( \frac{N!}{n_1! \cdots n_K!} \right) + \sum_{i=1}^K n_i \log p_i.
\end{equation}

Ignoring the constant term that does not depend on $\mathbf{p}$, the optimization problem reduces to:

\begin{align}
\max_{\mathbf{p}} \quad & \sum_{i=1}^K n_i \log p_i \\
\text{s.t.} \quad & \sum_{i=1}^K p_i = 1, \quad p_i \geq 0.
\end{align}

This is a standard constrained optimization problem, and the solution is obtained via the method of Lagrange multipliers. The resulting MLE estimator for each expert's activation probability is:

\begin{equation}
\hat{p}_i = \frac{n_i}{N}, \quad \forall i \in \{1, \ldots, K\}.
\end{equation}

\section{Predictor Architectures}
\label{sec:arch}

e formulate the expert selection problem in Mixture-of-Experts (MoE) as a multi-class classification task, where the objective is to predict the activated expert for each token in a sequence. Let $\mathcal{T}$ denote the set of input tokens, and let $E = \{e_1, \ldots, e_K\}$ denote the set of experts available in a given MoE layer. For each token $t \in \mathcal{T}$, the goal is to predict an expert label $y_t \in \{1, \ldots, K\}$ that will be used by the MoE routing mechanism.

We explore three modeling paradigms for this task: a global frequency-based model, a conditional frequency model, and neural network-based predictors.

\paragraph{Probability-Based Model.}
This baseline treats all tokens identically and assigns each token to the expert that is most frequently activated in the training data. Let $n_i$ be the number of times expert $e_i$ was selected across all tokens in the training corpus. The model estimates the global activation probabilities using maximum likelihood as:

\begin{equation}
\hat{p}_i = \frac{n_i}{\sum_{j=1}^{K} n_j}, \quad \forall i \in \{1, \ldots, K\}.
\end{equation}

The predicted expert for any token is then:

\begin{equation}
\hat{y}_t = \arg\max_{i} \hat{p}_i.
\end{equation}

This approach ignores token-specific context, providing a static prediction rule that reflects global expert utilization frequencies.

\paragraph{Conditional Probability Model.}
To improve over the static assignment, we consider a token- or position-conditioned frequency model. Let $I_t$ be the token index or its absolute position in the sequence. For each token index $i$, we count how many times each expert $e_k$ was selected and compute:

\begin{equation}
\hat{p}_{k \mid i} = \frac{n_{k,i}}{\sum_{j=1}^K n_{j,i}},
\end{equation}

where $n_{k,i}$ denotes the number of times token index $i$ selected expert $e_k$. The model then predicts:

\begin{equation}
\hat{y}_t = \arg\max_{k} \hat{p}_{k \mid I_t}.
\end{equation}

This conditional model captures per-token or per-position biases in expert activation.

\paragraph{Neural Networks.}
To learn token-aware expert selection strategies, we train neural models that take token embeddings as input and predict the corresponding expert activation for each MoE layer. Each model is trained with cross-entropy loss and optimized using the Adam optimizer. Input sequences are padded to a fixed length of 512 tokens during training, and separate classifiers are maintained for each layer of the MoE model.

We experiment with the following two architectures:

\begin{itemize}
  \item \textbf{Feed-Forward Network (FFN).} The FFN model is a lightweight two-layer MLP. Each input token embedding (of dimension 4096 for Mixtral) is first passed through a linear projection to a 128-dimensional hidden space, followed by a ReLU activation. This is then followed by another linear layer of the same hidden size. Finally, for each target MoE layer, a separate classifier head is implemented as a linear layer mapping from the 64-dimensional hidden state to 8 expert logits. The FFN model is shared across tokens and layers, with layer-specific output heads.
  
  \item \textbf{LSTM with Sparse Attention.} To capture temporal dependencies, we also design a recurrent model based on an LSTM encoder augmented with sparse attention. The input token embeddings are first projected from dimension 4096 (Mixtral) to 128 using a linear compression layer, followed by a ReLU activation. These projected embeddings are passed through a 2-layer LSTM with hidden size 64, applied in a batch-first manner. To enhance contextual modeling, we incorporate a sparse attention mechanism over the LSTM outputs (i.e., attention is applied using the LSTM outputs as query, key, and value). A residual connection is then added between the attention output and a separate feedforward transformation of the compressed input. Finally, for each MoE layer, a dedicated linear classifier maps the resulting vector to expert logits (8 classes for Mixtral).

\end{itemize}

\section{Results on LlaMA-MoE and Switch Transformer}
\label{sec:llama_and_switch_transformer}
\begin{figure}[t]
    \centering

    \begin{subfigure}[b]{0.33\textwidth}
        \centering
        \includegraphics[width=\linewidth]{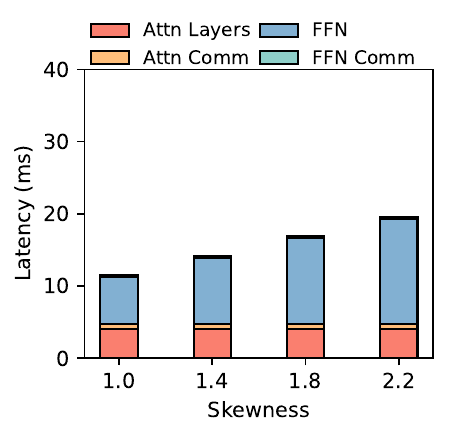}
        \caption{Baseline latency with no prediction (interconnect = NVLink).}
        \label{fig:llamamoe_prefill_skew_llamamoe_bs8_sl2048_gpu4_nvlink}
    \end{subfigure}%
    \hfill
    \begin{subfigure}[b]{0.63\textwidth}
        \centering
        \includegraphics[width=\linewidth]{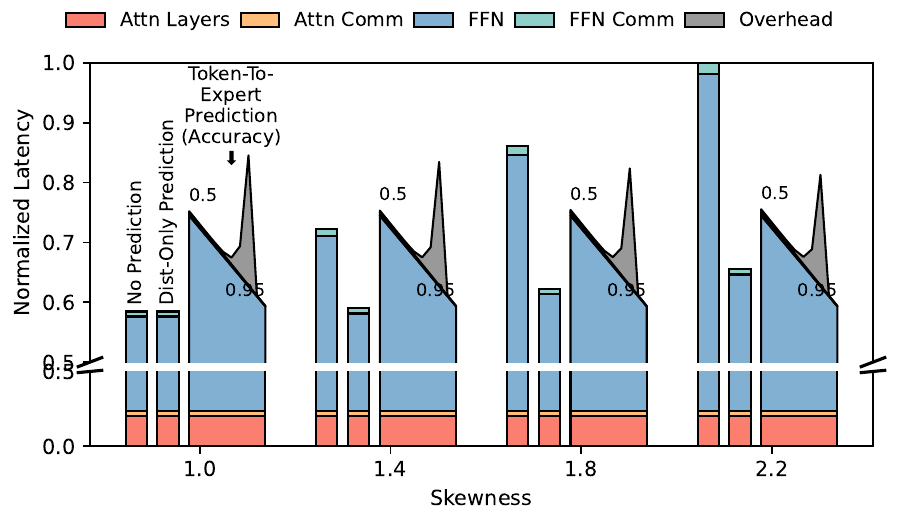}
        \caption{Latency of different prediction strategies and accuracies (interconnect = NVLink).}
        \label{fig:prefill_precise_prediction_llamamoe_bs8_sl2048_gpu4_nvlink}
    \end{subfigure}

    \vspace{0.7em}  

    \begin{subfigure}[b]{0.33\textwidth}
        \centering
        \includegraphics[width=\linewidth]{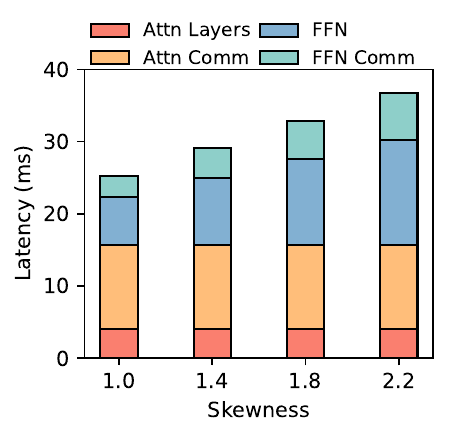}
        \caption{Baseline latency with no prediction (interconnect = PCIe).}
        \label{fig:llamamoe_prefill_skew_llamamoe_bs8_sl2048_gpu4_pcie}
    \end{subfigure}%
    \hfill
    \begin{subfigure}[b]{0.63\textwidth}
        \centering
        \includegraphics[width=\linewidth]{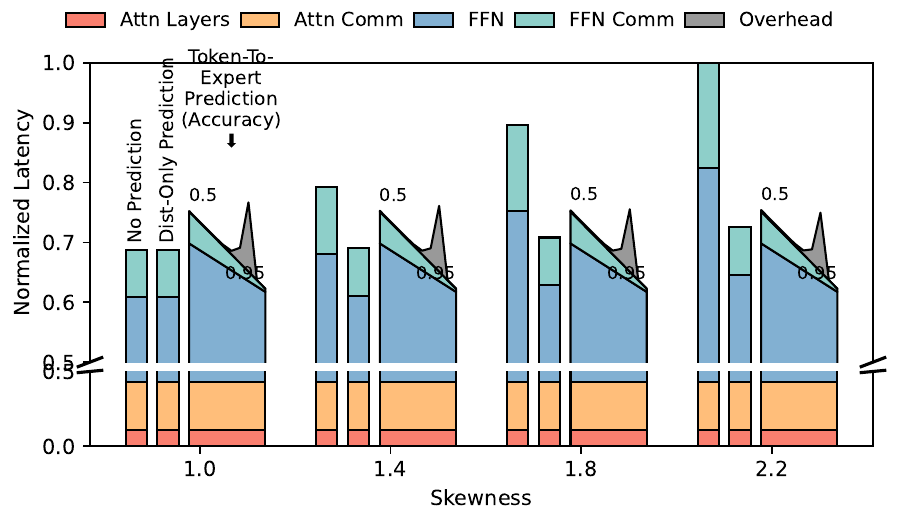}
        \caption{Latency of different prediction strategies and accuracies (interconnect = PCIe).}
        \label{fig:llamamoe_prefill_skew_llamamoe_bs8_sl2048_gpu4_pcie}
    \end{subfigure}
    \caption{Simulated prefill latency for a single layer of Llama-MoE model~\cite{zhu2024llama} under different prediction strategies and interconnect types. Workload sizes and hardware configurations are the same as Figure~\ref{fig:prefill_prediction_mixtral_full_comparison}. For illustration purposes, overhead > 0.5 of original latency is omitted. 
}
    \label{fig:prefill_prediction_llamamoe_full_comparison}
\vspace{-2em}
\end{figure}

To generalize our claim, we evaluated the performance implications and ran our simulations on other model architectures beyond Mixtral. We show results obtained from the Llama-MoE model~\cite{zhu2024llama} in Figure~\ref{fig:prefill_prediction_llamamoe_full_comparison} and the Switch Transformer model~\cite{fedus2022switch} in Figure~\ref{fig:prefill_prediction_switchtransformer_full_comparison}. We used the same datasets and the same hardware configurations as for the Mixtral experiments (MMLU, Alpaca Eval, and SST2; 4 A100 GPUs connected by NVLink or PCIe). 

Overall, the trends and the insights are similar to those we derived from the Mixtral model. We observed that the datasets generally have higher skewness in both models compared to Mixtral due to different routing decisions. We also noticed that it is more difficult to obtain very high prediction accuracy, and the prediction complexity required when approaching perfect prediction grows exponentially. For illustration purposes, we have omitted results where the overhead latency is greater than half of the original latency (layer-wise latency without overhead).

\begin{figure}[t]
    \centering

    \begin{subfigure}[b]{0.33\textwidth}
        \centering
        \includegraphics[width=\linewidth]{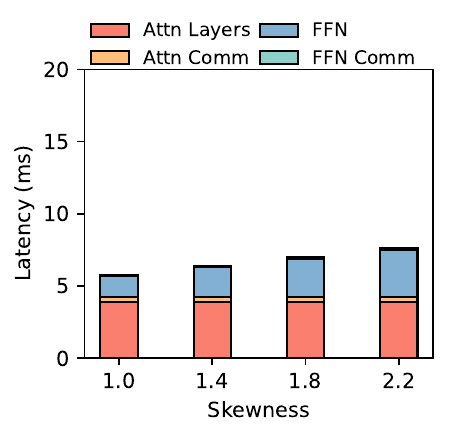}
        \caption{Baseline latency with no prediction (interconnect = NVLink).}
        \label{fig:switchtransformer_prefill_skew_switchtransformer_bs8_sl2048_gpu4_nvlink}
    \end{subfigure}%
    \hfill
    \begin{subfigure}[b]{0.63\textwidth}
        \centering
        \includegraphics[width=\linewidth]{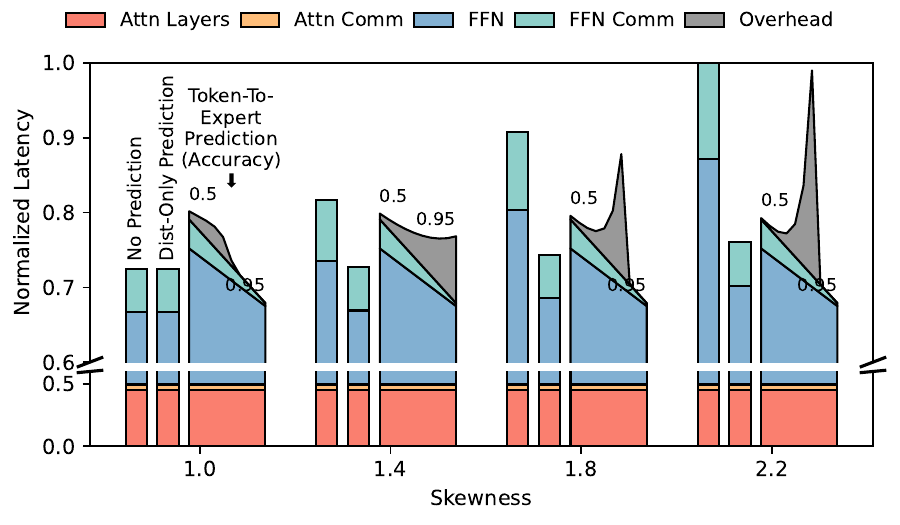}
        \caption{Latency of different prediction strategies and accuracies (interconnect = NVLink).}
        \label{fig:prefill_precise_prediction_switchtransformer_bs8_sl2048_gpu4_nvlink}
    \end{subfigure}

    \vspace{0.7em}  

    \begin{subfigure}[b]{0.33\textwidth}
        \centering
        \includegraphics[width=\linewidth]{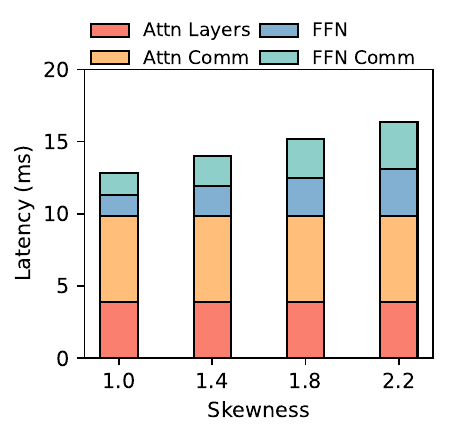}
        \caption{Baseline latency with no prediction (interconnect = PCIe).}
        \label{fig:switchtransformer_prefill_skew_switchtransformer_bs8_sl2048_gpu4_pcie}
    \end{subfigure}%
    \hfill
    \begin{subfigure}[b]{0.63\textwidth}
        \centering
        \includegraphics[width=\linewidth]{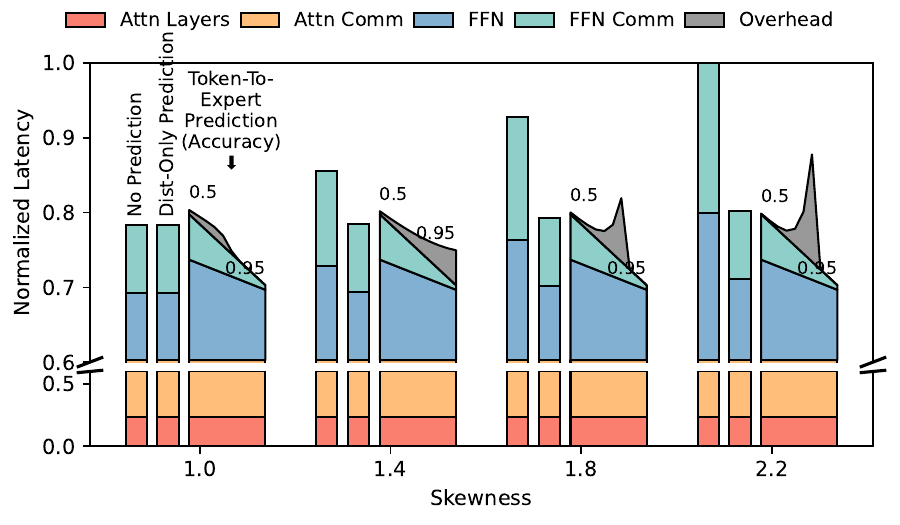}
        \caption{Latency of different prediction strategies and accuracies (interconnect = PCIe).}
        \label{fig:prefill_precise_prediction_switchtransformer_bs8_sl2048_gpu4_pcie}
    \end{subfigure}
    \caption{Simulated prefill latency for a single layer of Switch Transformer  model~\cite{fedus2022switch} under different prediction strategies and interconnect types. Workload sizes and hardware configurations are the same as Figure~\ref{fig:prefill_prediction_mixtral_full_comparison}. For illustration purposes, overhead > 0.5 of original latency is omitted. 
}
\end{figure}
    \label{fig:prefill_prediction_switchtransformer_full_comparison}
\vspace{-2em}
\newpage

\end{document}